%% file: paper.tex
\documentclass{article}
\usepackage[a4paper, total={6in, 8in}]{geometry}

\usepackage{microtype}
\usepackage{graphicx}
\usepackage{subfigure}
\usepackage{booktabs}

\usepackage{authblk}

\usepackage{biblatex}
\usepackage{hyperref}

\usepackage{amsmath}
\usepackage{amssymb}
\usepackage{mathtools}
\usepackage{amsthm}

\usepackage{makecell}

\usepackage[capitalize,noabbrev]{cleveref}

\usepackage[skip=10pt plus1pt, indent=0pt]{parskip}

\theoremstyle{plain}

\theoremstyle{definition}

\theoremstyle{remark}

\usepackage{tikz}
\usepackage{array}
\usepackage{multirow}
\usepackage{placeins}

\usetikzlibrary{bayesnet}
\usetikzlibrary{arrows}
\usetikzlibrary{backgrounds}

\DeclareMathOperator*{\argmin}{arg\,min}
\newcommand{\ourmodel}{GeDI}
\newcommand{\ouroperator}{\operatorname{GeDI}}

\author[1]{Luca Giuliani*}
\author[1]{Eleonora Misino*}
\author[1]{Michele Lombardi}
\affil[1]{Department of Computer Science and Engineering, University of Bologna, Bologna, Italy}
\date{}                   
\begin{document}

\title{Generalized Disparate Impact for Configurable Fairness Solutions in ML}

\maketitle
\def\thefootnote{*}\footnotetext{Equal contribution.\\
Correspondence to: Luca Giuliani luca.giuliani13@unibo.it, Eleonora Misino
eleonora.misino2@unibo.it}\def\thefootnote{\arabic{footnote}}

\begin{abstract}
We make two contributions in the field of AI fairness over continuous protected attributes.
First, we show that the Hirschfeld-Gebelein-Renyi (HGR) indicator (the only one currently available for such a case) is valuable but subject to a few crucial limitations regarding semantics, interpretability, and robustness.
Second, we introduce a family of indicators that are: 1) complementary to HGR in terms of semantics; 2) fully interpretable and transparent; 3) robust over finite samples; 4) configurable to suit specific applications.
Our approach also allows us to define fine-grained constraints to permit certain types of dependence and forbid others selectively.
By expanding the available options for continuous protected attributes, our approach represents a significant contribution to the area of fair artificial intelligence. 
\end{abstract}

\section{Introduction}
\input{src/intro.tex}

\section{Background and Motivation}
\input{src/background.tex}

\section{Generalized Disparate Impact}
\input{src/approach.tex}

\section{\ourmodel{} Computation and Constraints}
\input{src/computation.tex}


\section{Empirical Evaluation}
\input{src/experiments.tex}

\section{Conclusions}
\input{src/conclusions.tex}

\section*{Acknowledgements}
This work has been supported by the project TAILOR (funded by European Union’s Horizon 2020 research and innovation programme, GA No. 952215), and by the project AEQUITAS (funded by European Union's Horizon Europe research and innovation programme, GA No. 101070363)\footnote{Discaimer: This paper reflects only the authors’ views. The European Commission is not responsible for any use that may be made of the information it contains.}.

\printbibliography

\newpage
\appendix
\onecolumn
\input{src/appendix.tex}

\end{document}

%% file: src/intro.tex
In recent years, the social impact of data-driven AI systems and its ethical implications have become widely recognized.
For example, models may discriminate over population groups \cite{propublica2016,10.1001/jamainternmed.2018.3763}, spurring extensive research on AI fairness.
Typical approaches in this area involve quantitative indicators defined over a ``protected attribute'', which can be used to measure discrimination or enforce fairness constraints.
On the one hand, such metrics are arguably the most viable solution for mitigating fairness issues; on the other hand, the nuances of ethics can hardly be reduced to simple rules.
From this point of view, the availability of multiple and diverse metrics is a significant asset since it enables choosing the best indicator depending on the application.

Regarding available solutions, the case of categorical protected attributes is well covered by multiple indicators (see \Cref{sec:background}).
Conversely, a single approach works with continuous protected attributes at the moment of writing; this is the \emph{Hirschfeld-Gebelein-Renyi} (HGR) correlation coefficient \cite{Rnyi1959OnMO},  which has two viable implementations for Machine Learning (ML) systems \cite{pmlr-v97-mary19a,ijcai2020p313}.

We view the lack of diverse techniques for continuous protected groups as a major issue.
We contribute to this area by 1) identifying a few critical limitations in the HGR approach, and 2) introducing a family of indicators that complement HGR semantics and have technical advantages.

In terms of limitations, we highlight how the theoretical HGR formulation is prone to pathological behavior for finite samples, leading to the oversized importance of implementation details and limited interpretability.
Moreover, the generality of the HGR formulation makes the indicator unsuitable for exclusively measuring the functional dependency between the protected attribute and the target.
Finally, the HGR indicator cannot account for scale effects on fairness since it is based on the scale-invariant Pearson's correlation coefficient.

We introduce the Generalized Disparate Impact (\ourmodel), a family of indicators inspired by the HGR approach and by the Disparate Impact Discrimination Index \cite{DBLP:conf/aaai/AghaeiAV19}.
\ourmodel{} indicators measure the dependency based on how well a user-specified function of the protected attribute can approximate the target variable. 
Our indicators support both discrete and continuous protected attributes and 1) complement the HGR semantics, 2) are fully interpretable, 3) are robust for finite samples, and 4) can be extensively configured.
Indicators in the family share a core technical structure that allows for a uniform interpretation and unified techniques for stating fairness constraints. Moreover,  the \ourmodel{} formulation supports fine-grained constraints in order to permit only certain types of dependency while ruling out others.
By introducing \ourmodel{}, we aim to improve metrics diversity while limiting complexity.

%% file: src/background.tex
\label{sec:background}

State-of-the-art research on algorithmic fairness focuses on measuring discrimination and enforcing fairness constraints.
Some techniques \cite{DBLP:journals/kais/KamiranC11,NIPS2017_9a49a25d} operate by transforming the data before training the model to mitigate discrimination and are referred to as \emph{pre-processing approaches}. Conversely, \emph{post-processing approaches} calibrate the predictive model once trained \cite{Calders2010,NIPS2016_9d268236}. Finally, \emph{in-processing approaches} focus on removing discrimination at learning time.
To this end, \cite{pmlr-v81-menon18a} modify the class-probability estimates during training, while \cite{10.1007/978-3-642-33486-3_3,Zafar_2017,10.5555/3327144.3327203,ijcai2020p315,pmlr-v80-komiyama18a} embed fairness in the learning procedure through constraints in the objective.
All the approaches mentioned above rely on metrics restricted to discrete protected attributes.
Two recent works \cite{pmlr-v97-mary19a,ijcai2020p313} extend the method by \cite{6137441} to continuous variables by minimizing the Hirschfeld-Gebelein-Renyi (HGR) correlation coefficient \cite{Rnyi1959OnMO}.
A more extensive review of fairness approaches can be found in \cite{10.1145/3457607}.

\paragraph{Disparate Impact}
One of the most widely used notions of fairness is based on the legal concept of \textit{disparate impact} \cite{disparate_impact}, which occurs whenever a neutral practice negatively impacts a protected group. The principle of disparate impact can be extended to ML models by considering their output with respect to protected attributes \cite{DBLP:conf/kdd/FeldmanFMSV15}.
To quantify the disparate impact in regression and classification, \cite{DBLP:conf/aaai/AghaeiAV19} propose a fairness indicator deemed \textit{Disparate Impact Discrimination Index} (DIDI).
The higher the DIDI, the more disproportionate the model output is with respect to protected attributes, and the more it suffers from disparate impact. 

The simplicity of the DIDI formulation makes it highly interpretable.
Given a sample $\{x_i, y_i\}_{i=1}^n$ including values for a protected attribute $x$ and a continuous target value $y$, the \textit{Disparate Impact Discrimination Index} is referred to as $\mathrm{DIDI}_{\operatorname{r}}(x, y)$ and defined as:

\begin{equation}
    \label{eqn:didi_r}
    \sum_{v \in \mathcal{X}}\left|\frac{\sum_{i =1}^n y_i I(x_i=v)}{\sum_{i=1}^n I(x_i=v)}-\frac{1}{n} \sum_{i =1}^n y_i\right|
\end{equation}

where $\mathcal{X}$ is the domain of $x$ and $I(\phi)$ is the indicator function for the logical formula $\phi$.
The DIDI represents the sum of discrepancies between the average target for each protected group and for the whole dataset.
The indicator has a specialized formulation for classification tasks, i.e., when $y$ is discrete. In this case, $\mathrm{DIDI}_{\operatorname{c}}(x, y)$ is defined as:

\begin{equation*}
    \label{eqn:didi_c}
    \sum_{u \in \mathcal{Y}} \sum_{v \in \mathcal{X}} \left|\frac{\sum_{i=1}^n I(y_i=u \land x_i=v)}{\sum_{i=1}^n I(x_i=v)} - \frac{1}{n} \sum_{i=1}^n I(y_i=u)\right|
\end{equation*}

\paragraph{The HGR Indicator}
To the best of our knowledge, the HGR indicator is the only fairness metric that applies to continuous protected attributes. It is based on the \textit{Hirschfeld-Gebelein-Renyi} (HGR) correlation coefficient \cite{Rnyi1959OnMO}, which is a normalized measure of the relationship between two random variables, $X \in \mathcal{X}$ and $Y \in \mathcal{Y}$. When the coefficient is zero, the two variables are independent, while they are strictly dependent when it is equal to $1$. Formally, the HGR correlation coefficient is defined as:
\begin{equation}
    \operatorname{HGR}(X, Y)=\sup _{f, g} \rho(f(X), g(Y))
    \label{eq:hgr_def}
\end{equation}
where $\rho$ is Pearson's correlation coefficient and $f, g$ are two measurable functions (referred to as copula transformations) with finite and strictly positive variance.

\begin{figure*}[tb]
    \centering
     \subfigure[]{%
        \includegraphics[width=0.32\textwidth, height=0.2\textwidth]{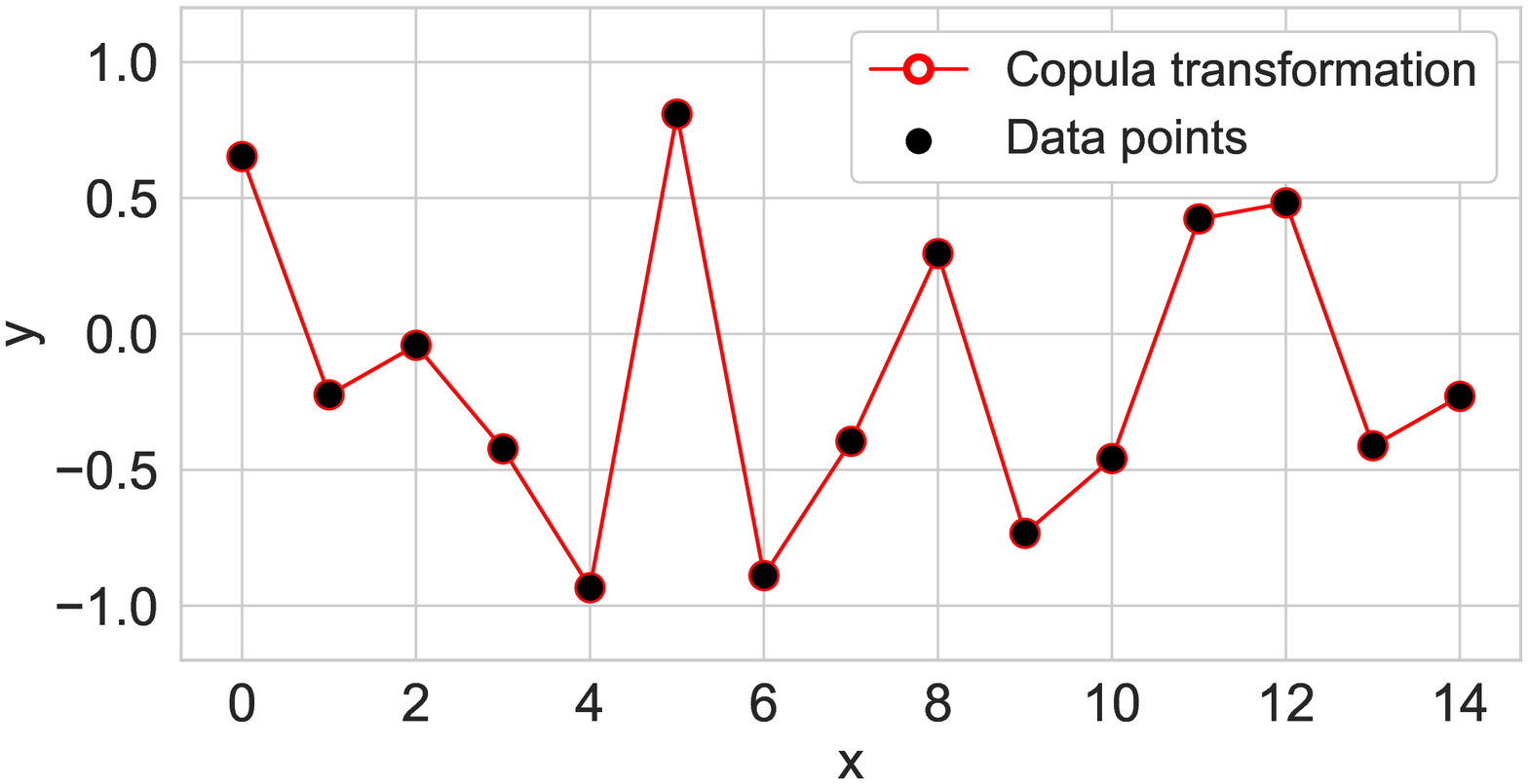}
        \label{fig:HGR_finite_sample}}
     \subfigure[]{
        \includegraphics[width=0.32\textwidth, height=0.2\textwidth]{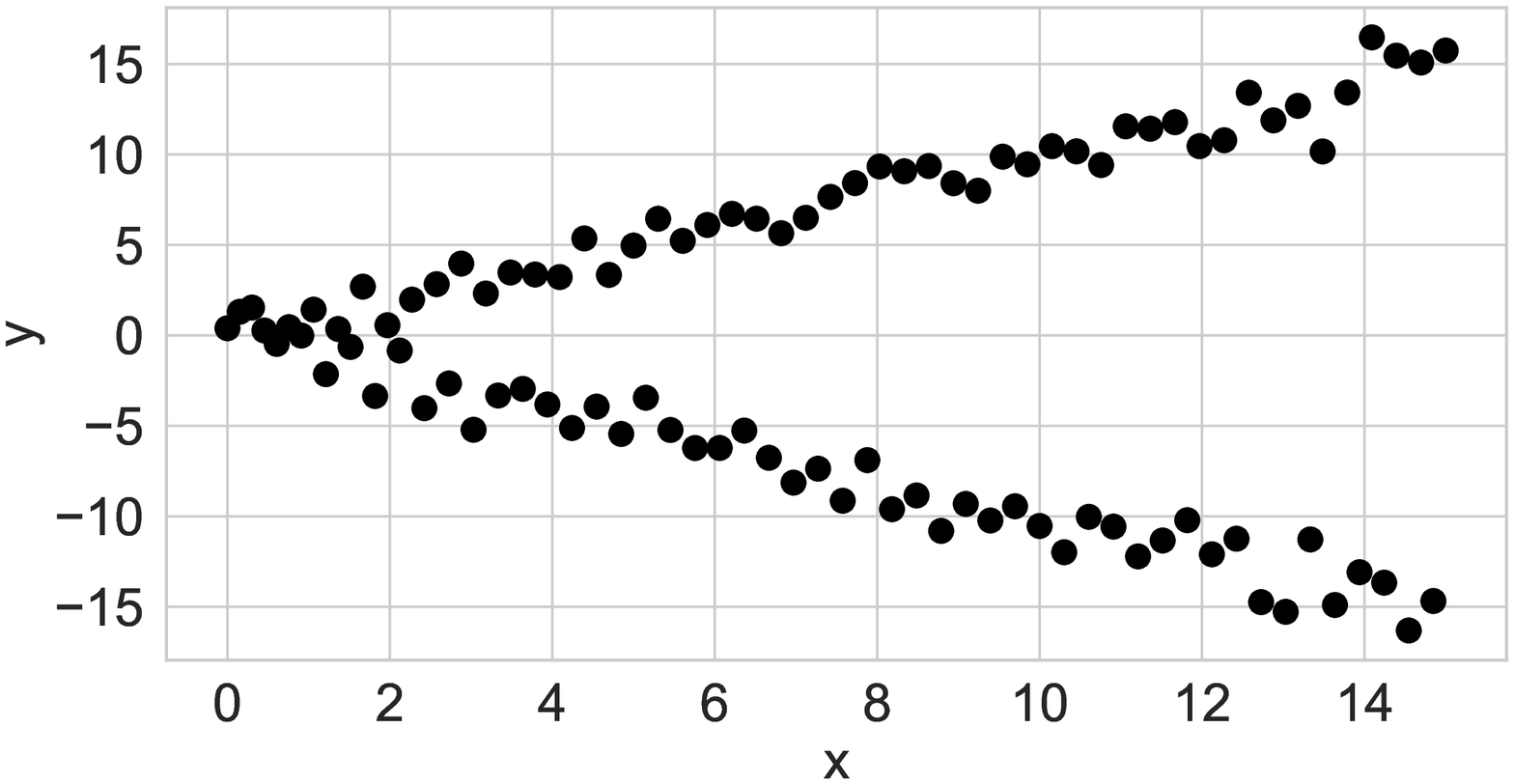}
        \label{fig:HGR_increasing_divergence}}
     \subfigure[]{
        \includegraphics[width=0.32\textwidth, height=0.2\textwidth]{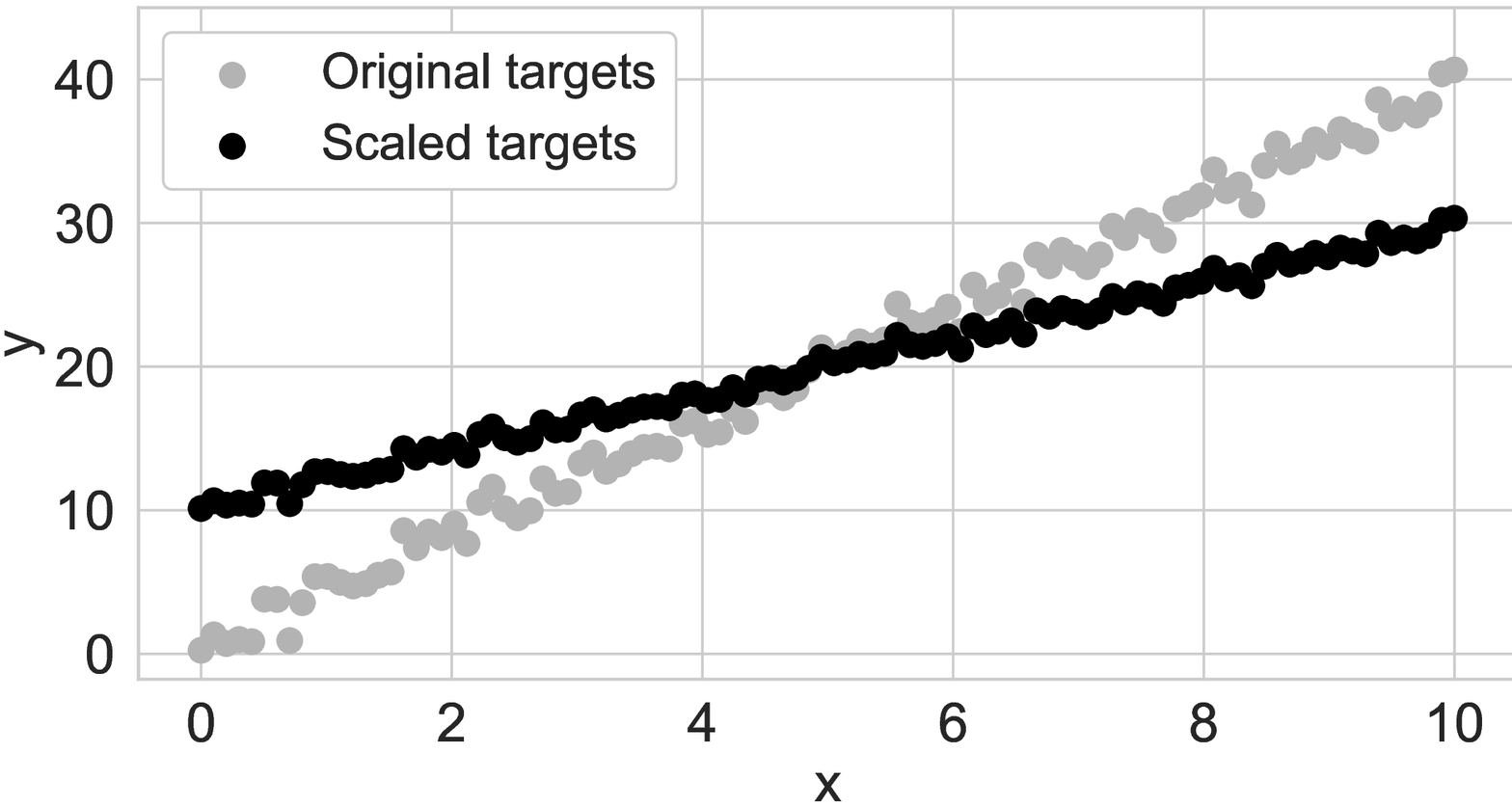}
        \label{fig:HGR_absolute_effectsB}}
    \caption{(a) HGR indicator may overfit data points due to the unrestricted transformations; this leads to overly large correlation estimates in finite datasets, even if $y$ is independent of $x$, as in the example. (b) Using unrestricted copula transformations allows HGR to capture non-functional dependencies, like the increasing divergence in target values. However, in some scenarios, discrimination is linked only to the function dependency $\mathbb{E}[y \mid x]$, and HGR is not able to measure it exclusively. (c) Since Pearson's correlation is scale-invariant, the HGR indicator is not able to capture scale effects on fairness.}

    \label{fig:HGR_limits}
\end{figure*}

\paragraph{Limitations of the HGR Approach}

The HGR indicator has several noteworthy properties, including the ability to measure very general forms of dependency.
Despite its strengths, however, it is not devoid of limitations. As a first contribution, we identify three of them.

First, when applied to finite samples, the theoretical HGR formulation is prone to pathological behavior.
In finite datasets, the Pearson's correlation in \Cref{eq:hgr_def} is replaced with its sample version, leading to:

\begin{equation}\label{eqn:hgr_sample}
    \operatorname{hgr}(x, y) = \max_{f, g} r(f_x, g_x)) = \max_{f, g} \frac{cov(f_x, g_y)}{\sigma(f_x)\sigma(g_y)}
\end{equation}
where $f_x , g_y,$ are short notations for $f(x)$ and $g(y)$, respectively; $\sigma(\cdot)$ is the sample standard deviation, and $cov(\cdot, \cdot)$ is the sample covariance.
Since the copula transformations are unrestricted, \Cref{eqn:hgr_sample} might be ill-behaved when the protected attribute takes many values. As an extreme case, with all-distinct $x$ values, choosing $f(x_i) = y_i$ ensures maximum Pearson's correlation, leading to overly large HGR values (see \Cref{fig:HGR_finite_sample}). The existing implementations mitigate this issue by using models with finite variance for $f$ and $g$ -- discretized KDE in \cite{pmlr-v97-mary19a}, and Neural Networks in \cite{ijcai2020p313}.
However, these solutions link the indicator semantics to low-level, sub-symbolic details that are difficult to interpret and to control.

Second, using unrestricted copula transformations allows HGR to measure very general forms of dependency between $x$ and $y$, including scenarios such as the diverging target values in \Cref{fig:HGR_increasing_divergence}.
However, there are situations where discrimination arises only when the \emph{expected} value of the target is affected, i.e., it is linked to the strength of the functional dependency $\mathbb{E}[y \mid x]$. For example, in \Cref{fig:HGR_increasing_divergence}, a third confounder attribute, correlated with $x$, might motivate the divergence. In this case, the confounder might not raise any ethical concern, but the HGR indicator would not able to exclusively measure the functional dependency.  

Third, the HGR indicator satisfies all the Renyi properties \cite{Rnyi1959OnMO} by relying on Pearson's correlation coefficient, but this makes the approach unable to account for scale effects on fairness.
For example, let us assume the target values $y$ are linearly correlated to the continuous protected attributes.
In some practical cases, applying an affine transformation on $y$ may reduce the discrimination (see \Cref{fig:HGR_absolute_effectsB}), but the HGR indicator cannot capture this effect since Pearson's correlation is scale-invariant.

%% file: src/approach.tex
\label{sec:approach}

In this section, we derive the \ourmodel{} family of indicators and its semantics. The process is guided by two design goals: 1) complementing the HGR approach to provide more options for continuous protected attributes; 2) improving over the technical limitations we identified in \Cref{sec:background}.

\paragraph{HGR Computation as Optimization}

We start by observing that the sample Pearson correlation can be restated as the optimal solution of a Least Squares problem. Formally, $r(f_x, g_y)$ from \Cref{eqn:hgr_sample} is given by:
\begin{equation}\label{eqn:pearson_as_optmization}
    \argmin_{r} \frac{1}{n} \left\|
    r \frac{f_x  - \mu(f_x)}{\sigma(f_x)}
    - \frac{g_y  - \mu(g_y)}{\sigma(g_y)}
    \right\|_2^2
\end{equation}
where $\mu(\cdot)$ is the sample average operator; this is a well-known statistical result, whose proof we report in Appendix~\ref{app:least_square_to_pearson}.
Using \Cref{eqn:pearson_as_optmization} may seem counter-intuitive since it casts the whole HGR computation process as a bilevel optimization problem. In particular, we have:
\begin{equation}
    \max_{f, g} \argmin_{r} \frac{1}{n} \left\|
    r \frac{f_x  - \mu(f_x)}{\sigma(f_x)}
    - \frac{g_y  - \mu(g_y)}{\sigma(g_y)}
    \right\|_2^2
\end{equation}

However, the two optimization objectives are aligned since larger $r$ values correspond to lower squared residuals; this alignment can be exploited to obtain an alternative single-level formulation for the HGR indicator.
The process is covered in detail in Appendix~\ref{app:bilevel_opt} and leads to:
\begin{equation}\label{eqn:hgr_alternative}
    \operatorname{hgr}(x, y) = |r^*|
\end{equation}
where $r^*$ is obtained by solving:
\begin{equation}\label{eqn:hgr_as_optimization}
    \argmin_{f, g, r} \frac{1}{n} \left\|
    r \frac{f_x  - \mu(f_x)}{\sigma(f_x)}
    - \frac{g_y  - \mu(g_y)}{\sigma(g_y)}  
    \right\|
\end{equation}
The equivalence holds if $\sigma(g_y),\sigma(f_x) > 0$, which is also needed for the Pearson correlation to be well-defined.

\paragraph{The \ourmodel{} Indicator Family}

The main insight from \Cref{eqn:hgr_alternative} and~\eqref{eqn:hgr_as_optimization} is that the HGR computation can be understood as a Least Square fitting.
We derive the \ourmodel{} family by building on the same observation, with a few key differences.
In particular, we define a \ourmodel{} indicator as a measure of how well a user-selected, interpretable function of the protected attribute $x$ can approximate the target variable $y$.
Formally, we have that:
\begin{equation}
    \label{eqn:indicator_abs_d}
    \ouroperator{}(x, y; F) = |d^*|
\end{equation}
where $d^*$ is defined via the optimization problem:
\begin{equation}\label{eqn:indicator_definition}
\begin{split}
    \argmin_{d, \alpha}\ & \frac{1}{n} \left\|
    d (F \alpha - \mu(F \alpha)) - (y - \mu(y))
    \right\|_2^2 \\
    \text{s.t. } & \|\alpha\|_1 = 1
\end{split}
\end{equation}
where $\alpha \in \mathbb{R}^k$ is a vector of coefficients, $d \in \mathbb{R}$ is a scale factor whose absolute value corresponds to the indicator value, and $\|\cdot\|_1$ is the L1 norm that we introduce to obtain one of the equivalent optimal solutions of the optimization problem.
$F$ is a $n \times k$ \emph{kernel matrix} whose columns  Finally, $F_j$ represent the evaluation of the basis functions $f_j(x)$ of the protected attribute $x$, namely:
\begin{equation}
    F \alpha = \sum_{j=1}^k F_j \alpha_j = \sum_{j=1}^k f_j(x) \alpha_j
\end{equation}
Any kernel can be chosen, provided that the resulting matrix is full-rank.  
The kernel and its order $k$ are part of the specification of a \ourmodel{} indicator and appear in its notation.

\begin{table}[bt]
\center
\caption{Type of dependencies measured by indicators for continuous protected attributes. Double check-mark ($\checkmark \checkmark$) specifies that the indicator \textit{exclusively} measures the corresponding dependency.
}
\vspace{3pt}
\begin{tabular}{lccc}
    \toprule
    \textbf{\small Measurable Dep.} &
    \textbf{\small \ourmodel{}} &
    \textbf{\small HGR-kde} &
    \textbf{\small HGR-nn} \\\midrule
    \small Functional
    & \small $\checkmark\checkmark$ & $\checkmark$ & $\checkmark$ \\
    \small Non-functional
    & $\times$ & $\checkmark$ & $\checkmark$ \\
    \small Scale-independent
    & $\checkmark$ & \small $\checkmark\checkmark$ & \small $\checkmark\checkmark$ \\
    \small Scale-dependent
    & $\checkmark$ & $\times$ & $\times$ \\
    \small User-configurable
    & $\checkmark$ & $\times$ & $\times$ \\\bottomrule
\end{tabular}

\label{tab:dep_types}
\end{table}

\paragraph{Rationale and Interpretation}

Our formulation differs from \Cref{eqn:hgr_as_optimization} in three regards.
First, it lacks the copula transformation on the $y$ variable (the $g$ function).
The DIDI inspires this property: it allows our indicators to exclusively measure the strength of the functional dependency $\mathbb{E}[y \mid x]$ but also makes them incapable of measuring other forms of dependency.
This semantics complements the HGR one, thus increasing the available options.

Second, the standardization terms have been replaced with the normalization constraint $\|\alpha\|_1 = 1$.
This constraint allows us to obtain one of the equivalent optimal solutions, while keeping it viable for linear optimization approaches.
The DIDI also inspires this choice as it makes our indicators sensitive to scale changes, complementing the HGR semantics.
As expected, this prevents the satisfaction of some Renyi properties that exclusively apply to scale-invariant metrics.

Third, we restrict the copula transformation on $x$ (the $f$ function) to be linear over a possibly non-linear kernel.
As a result, our indicator is fully interpretable. In particular, the indicator measures the overall functional dependency, while the coefficients identify which functional dependencies have the most significant effect, as determined by the kernel components.

\begin{figure*}[bt]
    \centering
    \subfigure[]{
        \includegraphics[width=0.22\linewidth]{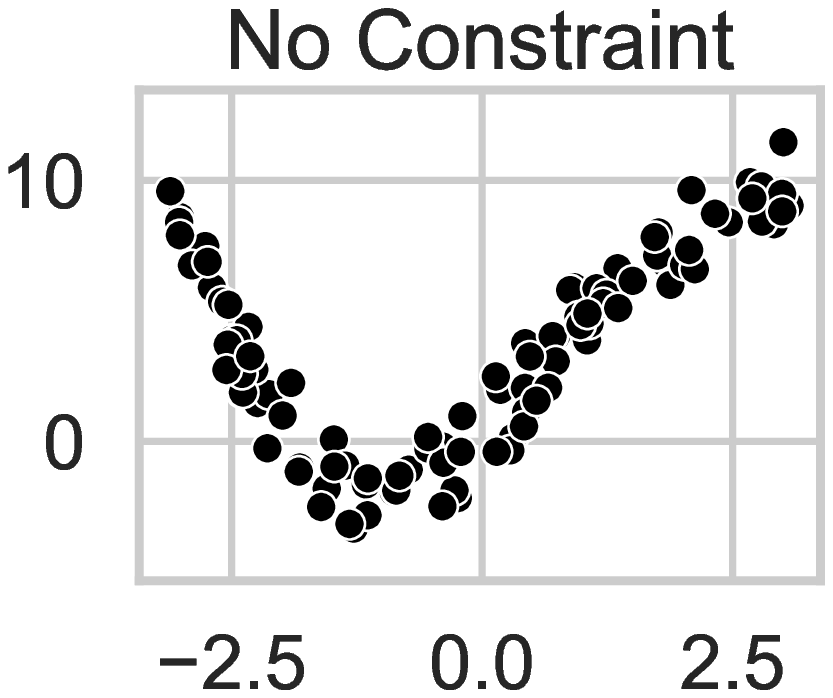}
        \label{fig:running_no}
    }
    \subfigure[]{
        \includegraphics[width=0.22\linewidth]{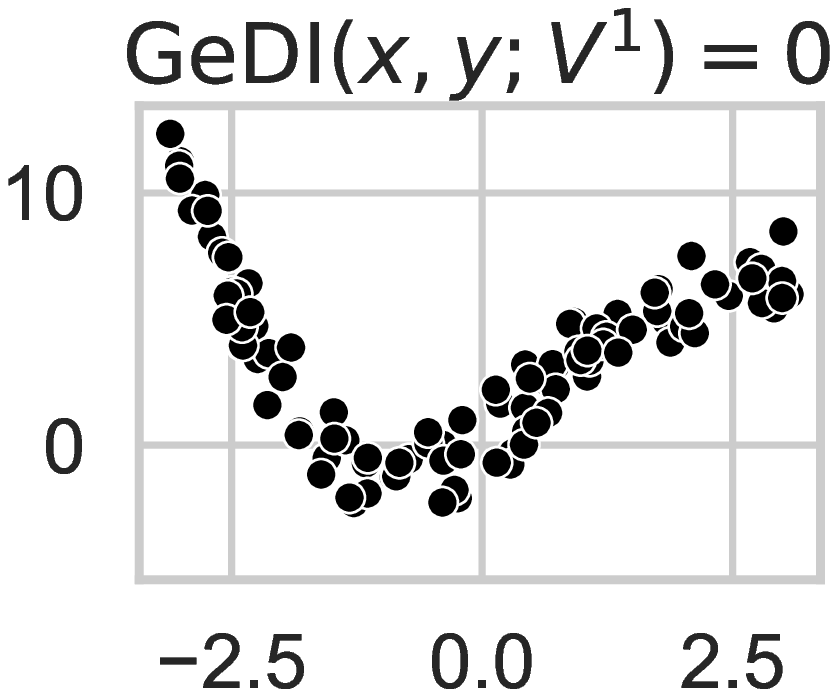}
        \label{fig:running_1}}
    \subfigure[]{
        \includegraphics[width=0.22\linewidth]{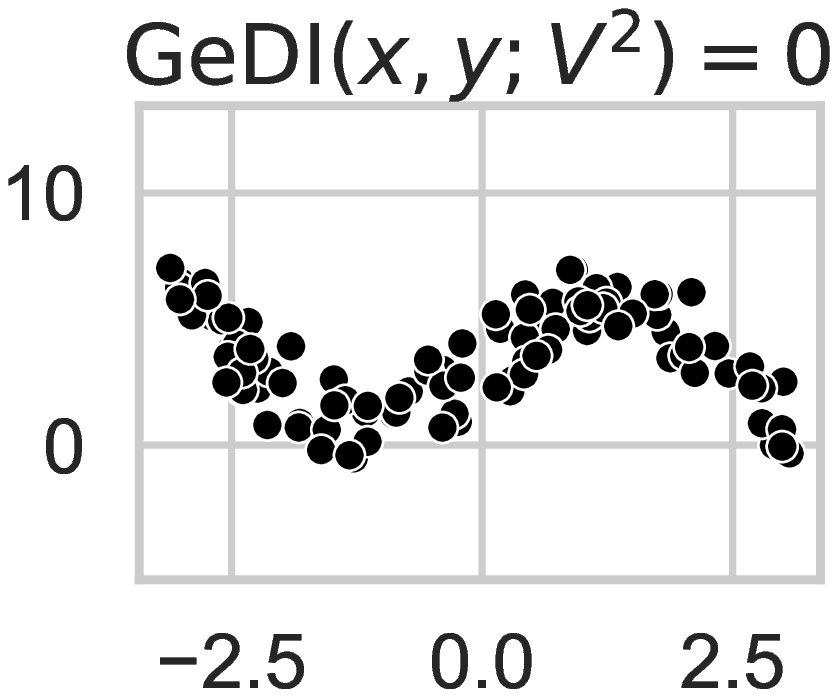}
        \label{fig:running_2}
    }
    \subfigure[]{
        \includegraphics[width=0.22\linewidth]{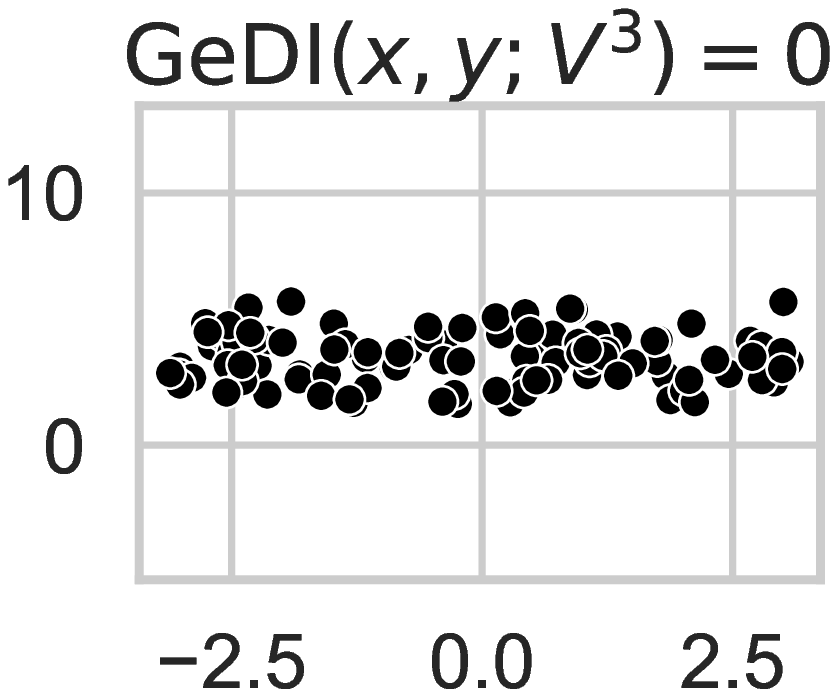}
        \label{fig:running_3}
    }
    \caption{Example of how the \ourmodel{} semantics depends on the kernel choice. Data in Figure 2(a) are synthetically generated by considering the relation  $y(x) = 4 sin(x) + x^2 + \epsilon$, where $\epsilon \sim \mathcal{N}(0, 1)$ and $x \sim \mathcal{U}(-\pi, \pi)$. The Figures 2(b)-(c)-(d) showcase the impact of increasing the kernel size in capturing the correlation between $x$ and $y$. More specifically, in 2(b) we apply a kernel of order 1, which is not able to capture most of the correlations, since the function has no linear terms; in 2(c) we use a kernel of order 2, which captures and cancels out the squared term in $y(x)$, preserving the sinusoidal component only; finally, in 2(d) we choose a kernel of order 3, which cancels out almost all of the sinusoidal component of $y(x)$ given that $sin(x)$ can be approximated as $x - \frac{1}{6} x^3$ as for the Taylor series, hence what is left is just $o(x^5)$ plus noise.}
    \label{fig:running_example}
\end{figure*}

\paragraph{Choice of Kernel}

Individual indicators from the \ourmodel{} family are obtained via the specification of a kernel, which allows users to define which types of functional dependency are relevant for the considered use case.
This is the main criterion for the kernel choice and provides a level of configurability that is absent in existing indicators.
\Cref{tab:dep_types} summarizes the types of dependency that can be measured by indicators with support for continuous protected attributes.
A double check-mark specifies that the indicator \emph{exclusively} measures the corresponding dependency.

Specifying a kernel might seem cumbersome, but it should be kept in mind that any HGR implementation also needs a mechanism to avoid ill behavior on finite samples.
In \cite{pmlr-v97-mary19a} this is done via a finite discretization and a KDE bandwidth, while in \cite{ijcai2020p313} via a neural network architecture and stochastic gradient descent.
In both cases, the mechanism is strongly linked to the implementation details, thus reducing transparency and control.
Integrating the kernel choice into the indicator definition makes the bias-variance trade-off explicit and controllable.

The kernel choice can be simplified by selecting a parametric function family and then adjusting the order $k$ to prevent overfitting and numerical issues.
If the chosen function family ensures the $F$ matrix is full-rank, we can asymptotically recover the unrestricted copula transformation $f$.
For example, by choosing a polynomial kernel, we have:
\begin{align}
    f_j(x) &= x^j
\end{align}
In our notation, $\ouroperator{}(\cdot, \cdot, V^k)$ denotes the use of a polynomial kernel of order $k$ ($V$ refers to the Vandermonde matrix).
Another suitable option is the Fourier kernel, which satisfies the full-rank property.
Both choices have a clear interpretation regarding either shape (for polynomials) or spectra (for Fourier kernels). \Cref{fig:running_example} shows how increasing the order of a polynomial kernel yields indicators sensitive to different types of dependence, thus requiring various adjustments with respect to a reference dataset to achieve an indicator value of 0. 
For example, $\ouroperator{}(x, y; V^2)$ is not sensitive to cubic dependence, which is measured by $\ouroperator{}(x, y; V^3)$.

Overall, we recommend the following process for selecting a kernel: 1) choose a family of functions based on relevance to the considered application and ease of interpretation; 2) tune the order to prevent overfitting and numerical issues.

\paragraph{\ourmodel{} and DIDI}

Our approach is designed so that, under specific circumstances, its behavior is analogous to the one of the DIDI \cite{DBLP:conf/aaai/AghaeiAV19}.
Let us start by differentiating the objective of \Cref{eqn:indicator_definition} in $d$, and stating the optimality condition (i.e., null derivative).
This yields:
\begin{equation}\label{eqn:indicator_covariance}
    \ouroperator{}(x, y; F) = \left|\frac{cov(F \alpha^*, y)}{var(F \alpha^*)}\right|
\end{equation}
whose full proof is in Appendix~\ref{app:gedi_computation}.
By assuming a polynomial kernel with $k = 1$ we have:
\begin{equation}\label{eqn:indicator_didi}
    \ouroperator{}(x, y; V^1) = \left|\frac{cov(x, y)}{var(x)}\right|
\end{equation}
where $\alpha^* = 1$, since there is a single coefficient and $\|\alpha^*\|_1 = 1$.
In Appendix~\ref{app:gedi_vs_didi}, we prove that \Cref{eqn:indicator_didi} is equivalent to the DIDI if the protected attribute is binary.
As a result, it is possible to consider the $\ouroperator{}(\cdot, \cdot; V^1)$ indicator as a generalization of the (binary) DIDI to continuous protected attributes, thus strengthening the link between our indicators and the already established metrics.

%% file: src/computation.tex
\label{sec:computation}

Next, we discuss how  \ourmodel{} indicators can be computed and used to enforce fairness constraints.

\paragraph{Computation}

Indicators in the \ourmodel{} family are defined via \Cref{eqn:indicator_definition}, which is a constrained optimization problem.
However, it is possible to obtain an unconstrained formulation by applying the following substitutions:
\begin{align}
    \tilde{F}_j &= F_j - \mu(F_j)
    & \tilde{y} &= y - \mu(y) 
    & \tilde{\alpha} &= |d| \alpha
    \label{eqn:tricky_substitution}
\end{align}
i.e., by centering $y$ and the columns $F_j$, and by combining $d$ and $\alpha$ in a single vector of variables.
The substitutions involve no loss of generality since \Cref{eqn:tricky_substitution} admits a solution for every $\tilde{\alpha} \in \mathbb{R}^k$.
As a result, the problem is reduced to 
a classical minimal Least Squares formulation:
\begin{equation}\label{eqn:indicator_unconstrained}
    \argmin_{\tilde{\alpha}} \frac{1}{n} \|
    \tilde{F} \tilde{\alpha} - \tilde{y}
    \|_2^2
\end{equation}
The solution can be computed via any Linear Regression approach, which typically involves solving the linear system:
\begin{equation}\label{eqn:unconstrained_solution_system}
    \tilde{F}^T \tilde{F} \tilde{\alpha}^* = \tilde{F}^T \tilde{y}
\end{equation}
From here, the absolute value $|d^*|$ can be recovered as:
\begin{equation}\label{eqn:convenient_definition}
    |d^*| = \|\tilde{\alpha}^*\|_1
\end{equation}
which is implied by \Cref{eqn:tricky_substitution} and the constraint $\|\alpha\|_1 = 1$.
Overall, the process is simple and well-understood in terms of numerical stability.
Additionally, since \Cref{eqn:indicator_unconstrained} can be solved to optimality in polynomial time, \ourmodel{} indicators are entirely determined by the kernel choice, with benefits for reproducibility.
This property is less clearly satisfied by the existing HGR-based indicators: the approach from  \cite{pmlr-v97-mary19a} is deterministic but ambiguous unless several implementation details are provided; the method from \cite{ijcai2020p313} is inherently not deterministic, since it relies on stochastic gradient descent.
In \Cref{tab:properties}, we summarize the characteristics of the \ourmodel{} indicators compared to the HGR approaches.

\begin{table}[t]
\center
\caption{Characteristics of the \ourmodel{} indicators when compared to HGR-kde \cite{pmlr-v97-mary19a} and HGR-nn \cite{ijcai2020p313} approach.}
\vspace{3pt}
\begin{tabular}{lccc}
    \toprule
    \textbf{\small Characteristic} &
    \textbf{\small \ourmodel{}} &
    \textbf{\small HGR-kde} &
    \textbf{\small HGR-nn} \\\midrule
    \small Interpretability
    & \small full & \small partial & \small partial\\[6pt]
    \makecell[l]{ \small Bias/variance trade-off}
    & \small transparent & \small opaque & \small opaque \\[6pt]
    \makecell[l]{\small Theoretically unrestricted \\ transformations}
    & $f$ & $f, g$ & $f, g$ \\[10pt]
    \small Renyi properties
    & \small partial & $\checkmark$ & $\checkmark$ \\[6pt]
    \small Deterministic
    & $\checkmark$ & $\checkmark$ & $\times$ \\[6pt]
    \makecell[l]{\small Information required \\ for full specification}
    & \small kernel & \makecell{\small discretization, \\ KDE bandwidth} & \small full NN + SGD params \\[10pt]
    \small Constraint enforcing
    & \small declarative \textbf{or} penalizer & \small penalizer & \small penalizer \\[6pt]
    \small Fine-grain constraints
    & $\checkmark$ & $\times$ & $\times$
    \\\bottomrule
\end{tabular}
\label{tab:properties}
\end{table}

\paragraph{Fairness Constraints}

\ourmodel{} indicators can be used to formulate fairness constraints. In this setting, the target $y$ can be changed, either because it represents the output of a ML model or because preprocessing/postprocessing techniques are employed.
A constraint in the form:
\begin{equation}
    \ouroperator{}(x, y; F) \leq q, \quad q \in \mathbb{R}^{+}
\end{equation}
can be translated into a set of piecewise linear relations:
\begin{align}
    & \tilde{F}^T \tilde{F} \tilde{\alpha}^* = \tilde{F}^T \tilde{y}
    \label{eqn:cst_set_1}\\
    & \|\tilde{\alpha}^*\|_1 \leq q
    \label{eqn:cst_set_2}
\end{align}
where $\tilde{\alpha}^*$ is not directly computed to avoid matrix inversion.
Specific projection-based approaches for constrained ML, such as the one from \cite{Detassis_2021}, can rely on mathematical programming to deal with constraints in this form.
Indeed, \Cref{fig:running_example} is obtained by adjusting target values to satisfy Equations \eqref{eqn:cst_set_1}-\eqref{eqn:cst_set_2} and minimizing the Mean Squared Error.
As an alternative, it is also possible to obtain a Lagrangian term (i.e., a penalizer) in the form:
\begin{equation}
    \label{eqn:gedi_vector_penalizer}
    \lambda \max(0, \|\tilde{\alpha}^*\|_1 - q), \quad \lambda \in \mathbb{R}^+
\end{equation}
where $\lambda$ is the penalizer weight. In order to avoid matrix inversion in automatic differentiation engines, we can rely on differentiable least-squares operators, such as \texttt{\small tf.linalg.lstsq} or \texttt{\small torch.linalg.lstsq}.

\paragraph{Fine-grained Constraints}
The choice of the kernel allows a user to specify which types of dependency should be measured. In addition, our formulation allows us to restrict specific terms of the copula transformation by enforcing constraints on individual coefficients.
Formally, it is possible to replace the set of constraints from \eqref{eqn:cst_set_2} with:
\begin{align}
    & |\tilde{\alpha}^*| \leq q, \quad q \in \mathbb{R}^{+k}
    \label{eqn:cst_set_2_fg}
\end{align}
where the single constraint on $\|\alpha^*\|_1$ is replaced by individual constraints on the absolute value of the coefficients.
By algebraic manipulation and taking advantage of the fact that $\tilde{F}^T \tilde{F}$ is positive definite, we obtain a single set of inequalities that avoids any matrix inversion:
\begin{align}\label{eqn:cst_set_3_fg}
    |\tilde{F}^T \tilde{y}|
    \leq 
    q \odot \tilde{F}^T \tilde{F}
\end{align}
where $\odot$ is the Hadamard (i.e., element-wise) product.
\Cref{eqn:cst_set_3_fg} can be processed directly by projection approaches or turned into a (vector) penalizer similar to \Cref{eqn:gedi_vector_penalizer}.

An interesting setup is to allow a single, easily explainable form of dependency (e.g.,  linear) while explicitly forbidding others; when applied to a ML model, this setup allows us to obtain a modified sample $\{x_i, y_i^\prime\}$, where $y_i^\prime$ represents the model output after the constraint has been enforced. In this situation, we have:
\begin{equation}
    \ouroperator{}(x, y^\prime; V^k) = \ouroperator{}(x, y^\prime; V^1)
\end{equation}
The equality holds since all polynomial degree dependencies from 2 to $k$ are explicitly removed, implying that we can measure the discrimination for the constrained model in terms of the DIDI analogy presented in \Cref{sec:approach}.

%% file: src/experiments.tex
\label{sec:experiments}

In this section, we discuss an empirical evaluation performed with three objectives: 1) testing how the kernel choice and fine-grained constraints affect the \ourmodel{} semantics; 2) studying the relation with other metrics, in particular the DIDI and the HGR indicators; 3) investigating how the use of different ML models and constraint enforcement algorithms impacts effectiveness.
Here we report the main details and experiments; additional information is provided in \Cref{app:section_5}. The source code and the datasets are available at \url{https://github.com/giuluck/GeneralizedDisparateImpact} under MIT license.

\begin{table}[bt]
    \center
    \caption{The four discrimination-aware learning tasks used in our experiments. Type B stands for binary, C for continuous.}
    \vspace{3pt}
    \small
    \begin{tabular}{m{5em} m{4.75em} >{\centering\arraybackslash}m{1.75em} m{4.75em} >{\centering\arraybackslash}m{1.75em}}
        \toprule
        &
        \multicolumn{2}{c}{\textbf{Target}} &
        \multicolumn{2}{c}{\textbf{Protected Att.}} \\
        \textbf{Dataset} &
        \textbf{Name} &
        \textbf{Type} &
        \textbf{Name} &
        \textbf{Type} \\
        \midrule
        \multirow{2}{*}{Adult} &
        \multirow{2}{*}{income} &
        \multirow{2}{*}{B} &
        sex &
        B \\
        & & &
        age &
        C \\ \noalign{\vskip 5pt}
        \multirow{2}{\textwidth}{Communities\\\& Crimes} &
        \multirow{2}{*}{violentPerPop} &
        \multirow{2}{*}{C} &
        race &
        B \\
        & & &
        pctBlack &
        C \\
        \bottomrule
    \end{tabular}
    \label{tab:experiments}
\end{table}

\paragraph{Experimental Setup}

\begin{figure}[h]
     \centering
     \begin{subfigure}[]{}
         \centering
         \includegraphics[width=0.7\textwidth]{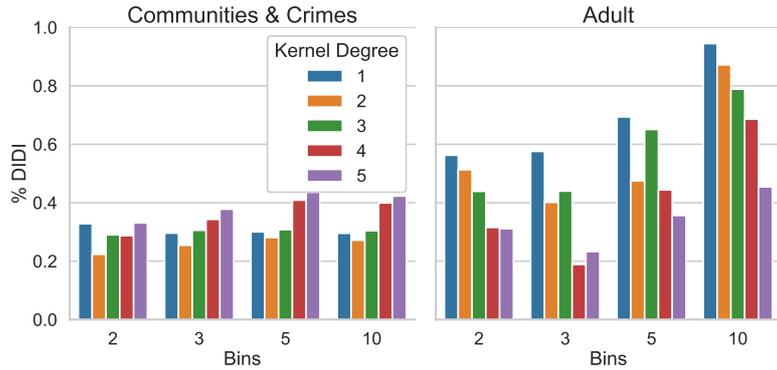}
        \label{fig:coarse_didi}
     \end{subfigure}%
    
    \centering
     \begin{subfigure}[]{}
         \centering
         \includegraphics[width=0.7\textwidth]{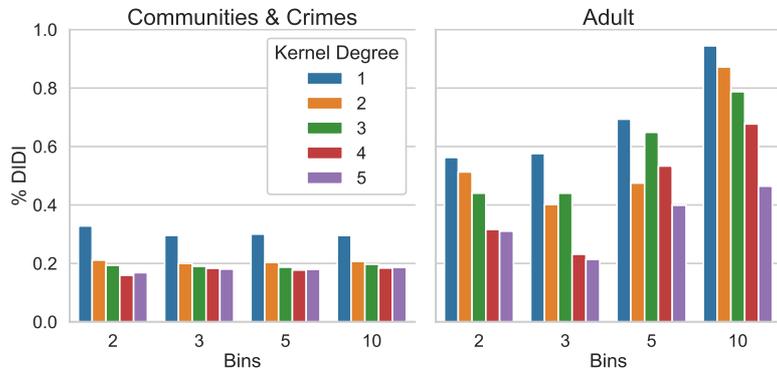}
        \label{fig:fine_didi}
     \end{subfigure}%

    \caption{Results from preprocessing tests presented from the perspective of the binned DIDI metrics with different bin numbers both for the coarse-grain (a) and the fine-grain (b) constraint formulation.}
    \label{fig:gedi_didi}
\end{figure}

We rely on two common benchmark datasets in the field of fair AI: \textit{Communities \& Crimes}\footnote{
\scriptsize{\url{https://archive.ics.uci.edu/ml/datasets/Communities+and+Crime}}}  with a continuous target, and \textit{Adult}\footnote{\scriptsize{\url{https://archive.ics.uci.edu/ml/datasets/Adult}}}  with a binary target.
For both, we define two discrimination-aware learning tasks by considering either a binary or a continuous protected attribute.
The resulting tasks are summarized in \Cref{tab:experiments}.

We restrict our analysis to polynomial kernels due to their ease of interpretation; moreover, their known issues with numerical stability at higher orders make them a computationally interesting benchmark.
We consider two setups for our method.
In the first, referred to as \ourmodel{}-Coarse, we derive an indicator via a $V^k$ kernel and constrain its value as in \Cref{eqn:cst_set_2}.
In the second, referred to as \ourmodel{}-Fine, we use the same type of indicator, but the constraint is enforced on individual coefficients; specifically, we allow some degrees of linear dependency while completely forbidding higher orders (up to $k$).
For binary protected attributes, we only use $k = 1$ as a higher order would not guarantee the full-rank requirement; hence the two setups coincide.

As a representative of the HGR approach, we use the indicator from \cite{pmlr-v97-mary19a} with default parameters.
For the DIDI, we use the original formulation from \cite{DBLP:conf/aaai/AghaeiAV19}.
When adopting DIDI for continuous protected attributes, we apply a quantile-based discretization in $n$ bins before computing the indicator; we refer to the resulting metrics as $\operatorname{DIDI}$-n. This technique does not require specialized indicators but is strongly sensitive to the chosen discretization and prone to overfitting with higher $n$ values.
We use \textit{percentage values} when presenting results for experiments with fairness constraints on all metrics, i.e., their values are normalized over those of the original datasets.

\paragraph{Experiments Description}

We consider two types of tests.
The first consist in \emph{preprocessing experiments}, where we directly change the target values to: 1) satisfy coarse- or fine-grained constraints on $\ouroperator{}(x, y; V^k)$ and 2) minimize a loss function appropriate to the task -- Mean Squared Error for regression and Hamming Distance for classification.
No machine learning model is trained in this process. We rely on an exact mathematical programming solver that allows for an efficient and stable formulation for fine-grained constraints (more details in \Cref{app:fine_grain_formulation}).
We use different kernel orders $k$, but a fixed threshold for each dataset, corresponding to 20\% of the value on the $\ouroperator{}(x, y, V^1)$ for the unaltered data.
With this design, any difference in behavior is entirely due to changes in the semantics determined by the chosen kernel.

Second, we report results for several \emph{performance tests}, where training problems with fairness constraints are solved using multiple machine learning models and optimization algorithms.
These experiments aim to test how these factors affect constraint satisfaction, model accuracy, and generalization.
In particular, we train an unconstrained version of a Random Forest (RF), a Gradient Boosting (GB), and a Neural Network (NN) model. 
Then, the same models are trained using the Moving Targets (MT) algorithm from \cite{Detassis_2021}, which allows us to deal with constraints in a declarative fashion.
Finally, to investigate the impact of different constraint enforcement methods, we train a version of the neural network model where a penalizer (or Semantics-Based Regularizer, SBR) is used during gradient descent. The penalizer weight is dynamically controlled at training time using the approach from \cite{Fioretto2021}. 
All these experiments are performed using a 5-fold cross-validation procedure on the entire dataset. See \Cref{app:experimental_setup} for a complete list of the technical settings, and  \Cref{app:constrained_models} for a broader description of the constrained models. Similarly to the preprocessing tests, we use a fixed threshold for all constraints, corresponding to 20\% the value of $\ouroperator{}(x, y, V^1)$ for the original datasets.

\input{tables/categorical.tex}

\input{tables/continuous.tex}

\paragraph{Preprocessing experiments}

\Cref{fig:gedi_didi} shows the outcome of our preprocessing tests for the $\ouroperator{}(x, y; V^k)$, both for coarse-grained (a) and fine-grained (b) approaches. In particular, we report the results for multiple kernel sizes (colors) and different bin numbers ($x$ axis) from the perspective of the binned DIDI metrics ($y$ axis).
All \ourmodel{} constraints are satisfied by construction due to the use of an exact solver.
Since increasing the bin number in DIDI-n enables measuring more complex dependencies, and both indicators focus on functional dependencies, the figures allow us to observe how the \ourmodel{} semantics depends on the kernel order.

We notice that increasing the kernel order with the \ourmodel{}-Fine approach tends to improve the discretized DIDI across all bin numbers (\Cref{fig:fine_didi}); in the few cases where the DIDI-3 and DIDI-5 worsen by adding \ourmodel{} constraints, we expect that the discrepancy is due to how the bin boundaries ``cut'' the shapes induced by the polynomial kernels, i.e., it is due to discretization noise.
The same behavior is observed for \ourmodel{}-Coarse (\Cref{fig:coarse_didi}), although with a lower degree of consistency.
Such a trend confirms that increasing the kernel order makes our indicators capable of measuring more complex dependencies.
In the \ourmodel{}-Fine case, all higher-order dependencies are canceled, while the \ourmodel{}-Coarse approach is more flexible regarding which types of dependency are allowed; this explains why enforcing a 20\% threshold with the \ourmodel{}-Fine approach also leads to similar DIDI-n values since the semantics of the two approaches are similar.
Despite this, we underline that the equivalence from \Cref{sec:computation} does not strictly hold in the analyzed datasets since the protected attribute is not natively categorical.

\paragraph{Performance experiments}

Table \ref{table:categorical} shows the results for the performance experiments in tasks with binary protected attributes.
We report the average values on the \emph{train} and \emph{validation} splits for: 1) an accuracy metric, i.e., \textit{R2} for regression tasks and \textit{Accuracy} for classification ones, and 2) our fairness indicator $\ouroperator(\cdot, \cdot; V^1)$, which in this case is equivalent to the DIDI.
Additionally, we report a single column with training times.
Rows related to unconstrained models are italicized, and the best results for the constrained models are highlighted in bold font.
Table \ref{table:continuous} reports similar data for tasks with continuous protected attributes. We rely on $\ouroperator{}(x, y, V^5)$ as a fairness indicator, and we enforce both \ourmodel{}-Fine and \ourmodel{}-Coarse since they differ in formulation for higher-order kernels.

The results obtained using different models and algorithms in the four tasks analyzed are rather diverse regarding accuracy, degree of constraint satisfaction, and training time.
This stresses the advantage of having access to multiple learning models and constraint enforcing methods, which our approach provides.
The behavior is reasonably consistent in terms of both accuracy and constraint satisfaction, with low standard deviations.
The only exception is the classification task (\emph{Adult}) with continuous protected attribute: in this scenario, and in general, when dealing with continuous protected attributes, the unconstrained models tend to introduce additional discrimination on top of the existing one.
As a consequence, our thresholds become particularly demanding since they are based on \ourmodel{} values for the original dataset.
Stringent thresholds require significant alterations to the input/output relation that the models would naturally learn, thus exacerbating generalization issues.

\begin{figure}[htp]
    \centering
    \includegraphics[width=0.48\textwidth]{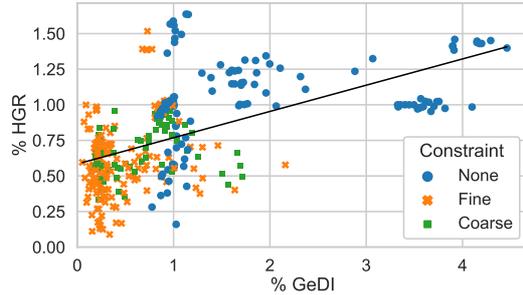}
    \caption{Percentage values of \ourmodel{} and HGR indicator from \cite{pmlr-v97-mary19a} on the performance tests.}
    \label{fig:gedi_hgr}
\end{figure}

Finally, in \Cref{fig:gedi_hgr}, we analyze the link between \ourmodel{} and the HGR indicator by reporting their values for all of our performance tests.
The displayed results come from unconstrained and constrained models and different tasks.
According to this figure, we highlight that: (1) GeDI is a valid option for fairness metrics, since it correlates with an already established metrics (namely HGR) but (2) it is not redundant, as its semantics differs from the one of HGR. Indeed, as expected, we can see that the two metrics are correlated (suggesting they capture related concepts of fairness), but with a high variance (highlighting the difference in their semantics). For this reason, GeDI offers a valid alternative to the set of existing metrics, enriching the range of options among which the practitioners can select the indicator that best fits their application.

%% file: tables/categorical.tex
\begin{table*}
\centering
\caption{Performance tests results for tasks with \textit{binary} protected feature.}
\label{table:categorical}
\vspace{2pt}
\scriptsize
\begin{tabular}{lccccl}
\toprule
{} & \multicolumn{5}{c}{\textit{Communities \& Crimes}}\\
{} & \multicolumn{2}{c}{R2} & \multicolumn{2}{c}{GeDI-V1 \%} & \multicolumn{1}{c}{Time}\\
{} & {train} & {val} & {train} & {val} & {}\\
\midrule
\it RF &\it 0.95 $\pm$ 0.00 &\it 0.63 $\pm$ 0.02 &\it 1.00 $\pm$ 0.00 &\it 1.02 $\pm$ 0.12 &\it 5.19 \\
~~+MT & 0.66 $\pm$ 0.01 & 0.52 $\pm$ 0.04 & 0.24 $\pm$ 0.00 & 0.28 $\pm$ 0.10 & 54.00\\[3pt]
\it GB &\it 0.87 $\pm$ 0.01 &\it 0.63 $\pm$ 0.02 &\it 1.03 $\pm$ 0.00 &\it 1.05 $\pm$ 0.11 &\it 2.43  \\
~~+MT & 0.64 $\pm$ 0.01 &\textbf{0.54} $\pm$ 0.02 & 0.21 $\pm$ 0.00 & 0.27 $\pm$ 0.09 & 29.25 \\[3pt]

\it NN &\it 0.99 $\pm$ 0.01 &\it 0.58 $\pm$ 0.01 &\it 0.99 $\pm$ 0.02 &\it 0.97 $\pm$ 0.12 &\it 7.64 \\

~~+MT &\textbf{0.89} $\pm$ 0.02 & 0.49 $\pm$ 0.04 & 0.20 $\pm$ 0.02 & 0.32 $\pm$ 0.11 & 85.81 \\

~~+SBR~~~~~~~ & 0.88 $\pm$ 0.01 & 0.45 $\pm$ 0.05 & \textbf{0.10} $\pm$ 0.03 & \textbf{0.24} $\pm$ 0.07 &\textbf{6.36} \\

\bottomrule
{} & \multicolumn{5}{c}{\textit{Adults}}\\
{} & \multicolumn{2}{c}{Accuracy} & \multicolumn{2}{c}{GeDI-V1 \%} & \multicolumn{1}{c}{Time}\\
{} & {train} & {val} & {train} & {val} \\
\midrule
\it RF &\it 1.00 $\pm$ 0.00 &\it 0.85 $\pm$ 0.00 &\it 1.00 $\pm$ 0.00 &\it 0.94 $\pm$ 0.05 &\it 1.54 \\
~~+MT & \textbf{0.95} $\pm$ 0.00 & 0.84 $\pm$ 0.00 & 0.20 $\pm$ 0.00 & 0.72 $\pm$ 0.03 &\textbf{128.28} \\[3pt]

\it GB & 0.87 $\pm$ 0.00 &\it 0.86 $\pm$ 0.00 &\it 0.88 $\pm$ 0.01 &\it 0.88 $\pm$ 0.04 &\it 2.25 \\
~~+MT & 0.85 $\pm$ 0.00 &\textbf{0.85} $\pm$ 0.00 & 0.35 $\pm$ 0.02 & 0.34 $\pm$ 0.04 & 129.65 \\[3pt]
\it NN &\it 0.89 $\pm$ 0.00 &\it 0.83 $\pm$ 0.00 &\it 1.03 $\pm$ 0.05 &\it 1.03 $\pm$ 0.08 &\it 55.31\\
~~+MT & 0.86 $\pm$ 0.00 & 0.83 $\pm$ 0.00 & 0.19 $\pm$ 0.01 & 0.17 $\pm$ 0.06 & 2593.11 \\
~~+SBR~~~~~~~ & 0.84 $\pm$ 0.00 & 0.84 $\pm$ 0.00 & \textbf{0.17} $\pm$ 0.02 & \textbf{0.17} $\pm$ 0.03 & 563.58 \\

\bottomrule
\end{tabular}
\end{table*}

%% file: tables/continuous.tex
\begin{table*}
\centering
\caption{Performance tests results for tasks with \textit{continuous} protected feature.}
\label{table:continuous}
\vspace{2pt}
\scriptsize
\begin{tabular}{lccccl}
\toprule
{} & \multicolumn{5}{c}{\textit{Communities \& Crimes}} \\
{} & \multicolumn{2}{c}{R2} & \multicolumn{2}{c}{GeDI-V5 \%} & \multicolumn{1}{c}{Time} \\
{} & {train} & {val} & {train} & {val} & {} \\
\midrule
\it RF &\it 0.95 $\pm$ 0.00 &\it 0.63 $\pm$ 0.02 &\it 1.70 $\pm$ 0.08 &\it 2.03 $\pm$ 0.59 &\it 5.23 \\
~~+MT-Fine & 0.48 $\pm$ 0.01 & 0.36 $\pm$ 0.02 & 0.24 $\pm$ 0.01 & 0.64 $\pm$ 0.20 & 61.97 \\
~~+MT-Coarse & 0.60 $\pm$ 0.01 & 0.46 $\pm$ 0.03 & 0.24 $\pm$ 0.01 & 0.69 $\pm$ 0.27 & 64.91  \\[4pt]
\it GB & 0.87 $\pm$ 0.01 &\it 0.63 $\pm$ 0.02 &\it 1.67 $\pm$ 0.10 &\it 2.09 $\pm$ 0.50 &\it 2.52  \\
~~+MT-Fine & 0.47 $\pm$ 0.01 & 0.37 $\pm$ 0.02 & 0.23 $\pm$ 0.00 & \textbf{0.55} $\pm$ 0.24 & 30.58  \\
~~+MT-Coarse & 0.60 $\pm$ 0.01 & \textbf{0.49} $\pm$ 0.04 & 0.22 $\pm$ 0.00 & 0.92 $\pm$ 0.44 & 31.93  \\[4pt]
\it NN &\it 0.99 $\pm$ 0.01 &\it 0.58 $\pm$ 0.01 &\it 1.79 $\pm$ 0.14 &\it 1.78 $\pm$ 0.35 &\it 7.58  \\
~~+MT-Fine & 0.70 $\pm$ 0.03 & 0.32 $\pm$ 0.05 & 0.27 $\pm$ 0.03 & 1.09 $\pm$ 0.67 & 94.14 \\
~~+MT-Coarse & \textbf{0.78} $\pm$ 0.06 & 0.40 $\pm$ 0.04 & 0.21 $\pm$ 0.02 & 0.90 $\pm$ 0.54 & 99.33 \\
~~+SBR-Fine & 0.64 $\pm$ 0.03 & 0.43 $\pm$ 0.04 & \textbf{0.18} $\pm$ 0.03 & 0.92 $\pm$ 0.47 & \textbf{15.87} \\
~~+SBR-Coarse & 0.67 $\pm$ 0.03 & 0.45 $\pm$ 0.02 & 0.21 $\pm$ 0.04 & 0.98 $\pm$ 0.52 & 16.20 \\

\bottomrule
{} & \multicolumn{5}{c}{\textit{Adults}} \\
{} & \multicolumn{2}{c}{Accuracy} & \multicolumn{2}{c}{GeDI-V5 \%} & \multicolumn{1}{c}{Time} \\
{} & {train} & {val} & {train} & {val} & {} \\
\midrule

\it RF &\it 1.00 $\pm$ 0.00 &\it 0.85 $\pm$ 0.00 &\it 3.40 $\pm$ 0.06 &\it 3.76 $\pm$ 0.18 &\it 2.74 \\
~~+MT-Fine  &\textbf{0.92} $\pm$ 0.00 & 0.78 $\pm$ 0.00 & \textbf{0.20} $\pm$ 0.00 & 1.13 $\pm$ 0.25 & 256.02 \\
~~+MT-Coarse & \textbf{0.92} $\pm$ 0.00 & 0.79 $\pm$ 0.00 & \textbf{0.20} $\pm$ 0.00 & 0.57 $\pm$ 0.19 & \textbf{60.89} \\[4pt]
\it GB &\it 0.87 $\pm$ 0.00 &\it 0.86 $\pm$ 0.00 &\it 3.59 $\pm$ 0.06 &\it 3.59 $\pm$ 0.16 &\it 4.14 \\
~~+MT-Fine & 0.81 $\pm$ 0.00 & 0.80 $\pm$ 0.01 & 0.65 $\pm$ 0.06 & 0.76 $\pm$ 0.26 & 658.27 \\
~~+MT-Coarse & 0.83 $\pm$ 0.00 & \textbf{0.83} $\pm$ 0.01 & 0.86 $\pm$ 0.03 & 0.90 $\pm$ 0.21 & 66.18 \\[4pt]
\it NN &\it 0.89 $\pm$ 0.00 &\it 0.83 $\pm$ 0.00 &\it 4.09 $\pm$ 0.16 &\it 4.12 $\pm$ 0.21 &\it 306.42 \\
~~+MT-Fine & 0.80 $\pm$ 0.01 & 0.79 $\pm$ 0.01 & 0.32 $\pm$ 0.08 & \textbf{0.51} $\pm$ 0.29 & 2096.00 \\
~~+MT-Coarse & 0.80 $\pm$ 0.01 & 0.78 $\pm$ 0.01 & 0.29 $\pm$ 0.05 & 0.63 $\pm$ 0.48 & 1378.40 \\
~~+SBR-Fine  & 0.84 $\pm$ 0.00 & \textbf{0.83} $\pm$ 0.00 & 0.89 $\pm$ 0.04 & 0.91 $\pm$ 0.10 & 793.28 \\
~~+SBR-Coarse & 0.84 $\pm$ 0.00 & \textbf{0.83} $\pm$ 0.00 & 0.91 $\pm$ 0.04 & 0.93 $\pm$ 0.06 & 925.15 \\
\bottomrule
\end{tabular}
\end{table*}

%% file: src/conclusions.tex
In this work, we introduce the \ourmodel{} family of indicators to extend the available options for measuring and enforcing fairness with continuous protected attributes.
Our indicators complement the existing HGR-based solutions regarding semantics and provide a more interpretable, transparent, and controllable fairness metric. \ourmodel{} allows the user to specify which types of dependency are relevant and how they should be restricted. 

While some of the design choices in our approach are incompatible with the HGR formulation, others could be applied in principle. The resulting configurable HGR-based solution would have technical properties similar to the \ourmodel{} indicators.
Preliminary work in this direction corroborates this conjecture, making us regard this topic as a promising area for future research.

\FloatBarrier

%% file: src/appendix.tex
\appendix
\onecolumn
\section{Proofs of Section 3}
\subsection{Least Squares Problem and Pearson's Correlation}
\label{app:least_square_to_pearson}
Consider two random samples $x \in \mathbb{R}^{n}$ and $y \in \mathbb{R}^{n}$ and the following least square problem:
\begin{equation}
    \argmin_{r} \frac{1}{n} \left\| r \frac{x - \mu_x}{\sigma_x} - \frac{y - \mu_y}{\sigma_y} \right\|_2^2
\end{equation}
where $r$ is the sample Pearson's correlation coefficient. This is a basic Linear Regression problem over standardized variables.
Since the problem is convex, an optimal solution can be found by requiring the loss function gradient to be null. By differentiating over $r$ we get:
\begin{equation}
    \frac{1}{n} \left(r \frac{x - \mu_x}{\sigma_x} - \frac{y - \mu_y}{\sigma_y}\right)^T \frac{x - \mu_x}{\sigma_x} = 0
\end{equation}
By algebraic manipulations we get:
\begin{equation}
    r \frac{1}{n} \frac{(x - \mu_x)^T (x - \mu_x)}{\sigma_x^2} = \frac{1}{n} \frac{(x - \mu_x)^T (y - \mu_y)}{\sigma_x \sigma_y}
\end{equation}

By applying the definition of variance and standard deviation, we have that $\frac{1}{n}(x - \mu_x)^T (x - \mu_x) = \sigma_x^2$, thus leading us to:
\begin{equation}\label{eqn:sample_rho}
    r = \frac{1}{n} \frac{(x - \mu_x)^T (y - \mu_y)}{\sigma_x \sigma_y}
\end{equation}
This is the value of the sample Pearson correlation coefficient, which is therefore equivalent to the optimal parameter for a properly defined Linear Regression problem.

\subsection{Bilevel Optimization Problem Simplification}
\label{app:bilevel_opt}
We prove that the objective of the outer and inner optimization problems are in this case aligned, thus making it possible to simplify the formulation.

Specifically, let us consider two pairs of copula transformations $f^{\prime}_x, g^{\prime}_y$ and $f^{\prime\prime}_x, g^{\prime\prime}_y$, and let $r^{\prime}$ and $r^{\prime\prime}$ be the corresponding values for the sample Pearson correlation coefficient. Let us assume that one pair of transformation leads to a smaller MSE, i.e.:
\begin{equation}\label{eqn:hgr_equiv_1}
    \frac{1}{n} \left\| r^{\prime} f^{\prime}_x - g^{\prime}_y \right\|_2^2 < \frac{1}{n} \left\| r^{\prime\prime} f^{\prime\prime}_x - g^{\prime\prime}_y \right\|_2^2
\end{equation}
The two terms can be expanded as:
\begin{equation}
    r^{\prime 2} \frac{f^{\prime T}_xf^{\prime}_x}{n} - 2 r^{\prime} \frac{f^{\prime T}_x g^{\prime}_y}{n} + \frac{g^{\prime T}_y g^{\prime}_y}{n}
    <
    r^{\prime \prime 2} \frac{f^{\prime\prime T}_x f^{\prime\prime}_x}{n} - 2 r^{\prime\prime} \frac{f^{\prime\prime T}_x g^{\prime\prime}_y}{n} + \frac{g^{\prime\prime T}_y g^{\prime\prime}_y}{n}
\end{equation}
Due to our zero-mean assumption, all quadratic terms in the form $f^{\prime T}_xf^{\prime}_x / n$, etc., correspond to sample variances. Again due to our assumptions, variances have unit value. Therefore, we have:
\begin{equation}
    r^{\prime 2} - 2 r^{\prime} \frac{f^{\prime T}_x g^{\prime}_y}{n} + 1
    <
    r^{\prime\prime 2} - 2 r^{\prime\prime} \frac{f^{\prime\prime T}_x g^{\prime\prime}_y}{n} + 1
\end{equation}
Now, as established in Equation~\eqref{eqn:sample_rho}, we have that $r^{\prime} = f^{\prime T}_x g^{\prime}_y / n$ and $r^{\prime\prime} = f^{\prime\prime T}_x g^{\prime\prime}_y / n$. Therefore, we have:
\begin{equation}\label{eqn:hgr_equiv_2}
    r^{\prime 2} - 2 r^{\prime 2} + 1
    <
    r^{\prime\prime 2} - 2 r^{\prime\prime 2} + 1
\end{equation}
Since all steps leading from Equation~\eqref{eqn:hgr_equiv_1} to Equation~\eqref{eqn:hgr_equiv_2} are invertible, we have that:
\begin{equation}\label{eqn:hgr_equiv}
    r^{\prime 2} > r^{\prime\prime 2}
    \quad \Leftrightarrow \quad
    \frac{1}{n} \left\| r^{\prime} f^{\prime}_x - g^{\prime}_y \right\|_2^2 < \frac{1}{n} \left\| r^{\prime\prime} f^{\prime\prime}_x - g^{\prime\prime}_y \right\|_2^2
\end{equation}
In other words, maximizing the square of the sample HGR is equivalent to minimizing the Mean Squared Error.
Now, maximizing $r^2$ corresponds to maximizing either $r$ or minimizing $-r$. Since the copula transformations are generic, we can always change the sign of $r$ by changing the sign of either $f$ or $g$. This means that maximizing $r$ is in fact equivalent to maximizing $r^2$ in our context.

Overall, this suggests an approach for computing the sample HGR that does not rely on bilevel optimization. Namely, first we determine $f$ and $g$ by solving the Least Squares Problem:
\begin{equation}
    \argmin_{f, g, r} \frac{1}{n} \left\| r f_x - g_y \right\|_2^2
\end{equation}
Then we compute the sample HGR as the absolute value of the sample Pearson correlation, i.e.:
\begin{equation}
    \operatorname{hgr}(x, y) = |r(f_x, g_x)|
\end{equation}
Computing the absolute value is equivalent to performing the sign swap on one of the two transformations, as previously described.

\subsection{Closed-form Computation of Generalized Disparate Impact}
\label{app:gedi_computation}

We start from the definition of the GeDI family of indicators as for \Cref{eqn:indicator_definition}, i.e.:

\begin{align}
    \argmin_{d, \alpha}\ \frac{1}{n} \left\|
    d (F \alpha - \mu(F \alpha)) - (y - \mu(y))
    \right\|_2^2 & & \text{s.t. } \|\alpha\|_1 = 1
\end{align}

We can embed the constraint into the objective function $C(d, \alpha, \lambda)$ using a Lagrangian multiplier $\lambda$, from which we obtain:

\begin{align}
    \argmin_{d, \alpha, \lambda}\ C(d, \alpha, \lambda) & & \text{s.t. } C(d, \alpha, \lambda) = \frac{1}{n} \left\| d (F \alpha - \mu(F \alpha)) - (y - \mu(y))
    \right\|_2^2 + \lambda (\|\alpha\|_1 - 1)
\end{align}

The optimal solution of the objective function can be found by requiring its gradient to be null. This implies that having a null derivative in $d$ is a necessary condition for optimality, which we can exploit to retrieve the value of $d^*$ as follows. First, we compute the derivative of $C(d, \alpha, \lambda)$ with respect to $d$:

\begin{equation}
    \label{eqn:gedi_derivative}
    \frac{\partial C(d, \alpha, \lambda)}{\partial d} = \frac{2}{n} (F \alpha - \mu(F \alpha))^T ((F \alpha - \mu(F \alpha))d - (y - \mu(y)))   
\end{equation}

Then, by requiring \Cref{eqn:gedi_derivative} to be null, we get the following equivalence:
\begin{equation}
    \frac{1}{n} (F \alpha^* - \mu(F \alpha^*))^T (F \alpha^* - \mu(F \alpha^*)) d^* = \frac{1}{n} (F \alpha^* - \mu(F \alpha^*))^T (y - \mu(y))
\end{equation}

The scalar values $\frac{1}{n} (F \alpha^* - \mu(F \alpha^*))^T (F \alpha^* - \mu(F \alpha^*))$ and $\frac{1}{n} (F \alpha^* - \mu(F \alpha^*))^T (y - \mu(y))$ represent the variance of the vector $F \alpha^*$ and the covariance between $F \alpha^*$ and $y$, respectively. Hence, we get the closed-form value of $d^*$ as:
\begin{equation}
    d^* = \frac{\operatorname{cov}(F \alpha^*, y)}{\operatorname{var}(F \alpha^*)}
\end{equation}

Finally, since \Cref{eqn:indicator_abs_d} ties the value of the \ourmodel{} indicator to the absolute value of $d^*$, we get:

\begin{equation}
    \ouroperator{}(x, y; F) = \left| \frac{\operatorname{cov}(F \alpha^*, y)}{\operatorname{var}(F \alpha^*)} \right|
\end{equation}

\subsection{Equivalence Between GeDI and DIDI}
\label{app:gedi_vs_didi}

We consider the case of continuous target $y$ and binary protected attribute $x$. The $\operatorname{DIDI}$ can be computed as for \Cref{eqn:didi_r}:

\begin{equation}
    \operatorname{DIDI}_r(x, y) = \sum_{v \in \mathcal{X}}\left|\frac{\sum_{i =1}^n y_i I(x_i=v)}{\sum_{i=1}^n I(x_i=v)} - \frac{1}{n} \sum_{i =1}^n y_i\right|
\end{equation}

Since $x$ is a binary attribute ($\mathcal{X} = \{0, 1\}$), we can replace the indicator function $I(x_i = v)$ with either $1 - x_i$ or $x_i$ depending on the value $v$, obtaining:

\begin{equation}
    \label{eqn:didi_special_regression}
    \operatorname{DIDI}_r(x, y) = \left| \frac{\sum_{i =1}^n (1 - x_i) y_i}{\sum_{i=1}^n (1 - x_i)} - \frac{1}{n} \sum_{i =1}^n y_i \right| + \left| \frac{\sum_{i =1}^n x_i y_i}{\sum_{i=1}^n x_i} - \frac{1}{n} \sum_{i =1}^n y_i \right|
\end{equation}

By algebraic manipulations within the summations, and by dividing for the constant factor $n$ both the numerator and the denominator of every fractional term, we can rewrite \Cref{eqn:didi_special_regression} as:
\begin{equation}
    \label{eqn:didi_special_means_1}
    \operatorname{DIDI}_r(x, y) = \left| \frac{\mu_y - \mu_{xy}}{1 - \mu_x} - \mu_y \right| + \left| \frac{\mu_{xy}}{\mu_x} - \mu_y \right|
\end{equation}
where $\mu_x$ and $\mu_y$ represent the average value of vectors $x$ and $y$ respectively, and $\mu_{xy}$ is the average value of vector $x \odot y$.

We can further manipulate this equation to obtain:
\begin{equation}
    \label{eqn:didi_special_means_2}
    \operatorname{DIDI}_r(x, y) = \left| \frac{\mu_x\mu_y - \mu_{xy}}{1 - \mu_x}\right| + \left| \frac{\mu_{xy} - \mu_x\mu_y}{\mu_x} \right|
\end{equation}

The numerators of both terms are equal except for the sign, hence we can join the two absolute values by simply swapping the sign of one of them. Moreover, we can notice that $\mu_{xy} - \mu_x \mu_y$ represents the covariance between $x$ and $y$. Therefore:
\begin{equation}
    \label{eqn:didi_special_covariance}
    \operatorname{DIDI}_r(x, y) = \left| \frac{\operatorname{cov}(x, y)}{1 - \mu_x} + \frac{\operatorname{cov}(x, y)}{\mu_x} \right| = \left| \frac{\operatorname{cov}(x, y)}{\mu_x - \mu_x^2} \right|
\end{equation}

Since $x$ is a binary vector, it is invariant to the power operator. Thus,  $\mu_x = \mu_{x^2}$ and, subsequently, the denominator of \Cref{eqn:didi_special_covariance} reduces to the variance of $x$. Hence:
\begin{equation}
    \operatorname{DIDI}_r(x, y) = \left| \frac{\operatorname{cov}(x, y)}{\operatorname{var(x)}} \right| = \ouroperator{}(x, y; V^1)
\end{equation}

The same reasoning can be applied when both $x$ and $y$ are binary vectors. Again, the indicator function $I(x_i = v)$ can be replaced with either $1 - x_i$ or $x_i$ depending on the value $v$. Similarly, since in this case we are computing the $\operatorname{DIDI}_c$, we replace $I(y_i = u)$ with $1 - y_i$ and $y_i$, and $I(y_i = u \land x_i = v)$ with the corresponding product between the previous terms. Eventually, we obtain:
\begin{equation}
\begin{split}
    \label{eqn:didi_special_classification}
    \operatorname{DIDI}_c(x, y) = \left| \frac{\sum_{i =1}^n (1 - x_i) (1 - y_i)}{\sum_{i=1}^n (1 - x_i)} - \frac{1}{n} \sum_{i =1}^n (1 - y_i) \right| + \left| \frac{\sum_{i =1}^n x_i (1 - y_i)}{\sum_{i=1}^n x_i} - \frac{1}{n} \sum_{i =1}^n (1 - y_i) \right| + \\
    \left| \frac{\sum_{i =1}^n (1 - x_i) y_i}{\sum_{i=1}^n (1 - x_i)} - \frac{1}{n} \sum_{i =1}^n y_i \right| + \left| \frac{\sum_{i =1}^n x_i y_i}{\sum_{i=1}^n x_i} - \frac{1}{n} \sum_{i =1}^n y_i \right|
\end{split}
\end{equation}

By using the same notation as in \Cref{eqn:didi_special_means_1} and applying analogous mathematical manipulations, we get to:
\begin{equation}
    \operatorname{DIDI}_c(x, y) = \left| \frac{\mu_x\mu_y - \mu_{xy}}{1 - \mu_x}\right| + \left| \frac{\mu_{xy} - \mu_x\mu_y}{\mu_x} \right| + \left| \frac{\mu_x\mu_y - \mu_{xy}}{1 - \mu_x}\right| + \left| \frac{\mu_{xy} - \mu_x\mu_y}{\mu_x} \right|
\end{equation}

This value is twice as much that in \Cref{eqn:didi_special_means_2}, meaning that the $\operatorname{DIDI}_c$ is twice our indicator $\ouroperator{}(x, y; V^1)$. This is not a problem since we can get rid of the constant scaling factor. Moreover, when we constrain the DIDI indicator up to a relative threshold that depends on the original level of discrimination, the scaling factor cancels out naturally.

\section{Appendix of Section 5}
\label{app:section_5}

\subsection{Optimized Fine-grained Formulation}
\label{app:fine_grain_formulation}

We want to solve the following optimization model:
\begin{equation}
\begin{split}
    \label{eqn:fine_grained_problem}
    & \argmin_z \left\| z - y \right\|_2^2 \\
    \text{s.t. } \ouroperator{}&(x, z; V^k) = \ouroperator{}(x, z; V^1), \\
    \ouroperator{}&(x, z; V^1) \leq q
\end{split}
\end{equation}
where the \ourmodel{} indicator is defined as in \Cref{eqn:indicator_abs_d} and computed according to Equations \eqref{eqn:unconstrained_solution_system}-\eqref{eqn:convenient_definition}.

Since we impose $\ouroperator{}(x, z; V^k) = \ouroperator{}(x, z; V^1)$, the optimal vector $\tilde\alpha^*$ which solves \Cref{eqn:unconstrained_solution_system} for $F = V^k$ is such that all the elements are null apart from the first one.
It follows that:
\begin{equation}
    \label{eqn:fine_grained_system}
    \tilde{F_1}^T \tilde{F} \tilde\alpha_1^* = \tilde{F}^T \tilde{y}
\end{equation}
where $\tilde{F}_1 = x - \mu(x) = \tilde{x}$ is the first column of the kernel matrix, namely the only one paired to a non-null $\tilde\alpha$ coefficient.

\Cref{eqn:fine_grained_system} is a system of $k$ equations with a single variable. From the first equation we can retrieve the value of $\tilde\alpha_1^*$ in the following way:
\begin{equation}
    \tilde{x}^T \tilde{x} \tilde\alpha_1^* = \tilde{x}^T \tilde{y}
\end{equation}
whose solution is $\tilde\alpha_1^* = \frac{\operatorname{cov}(x, y)}{\operatorname{var}(x)}$.
According to \Cref{eqn:convenient_definition}, the \ourmodel{} indicator can be retrieved as the absolute value of $\tilde\alpha_1^*$ since all the remaining items of $\tilde\alpha^*$ are null. This result is in fact equivalent to $\ouroperator{}(x, y; V^1)$.

In addition to that, the remaining $k - 1$ equalities from \Cref{eqn:fine_grained_system} must be satisfied. This can be achieved by operating on the projections. Indeed, the optimization problem defined in \Cref{eqn:fine_grained_problem} has $n$ free variables -- i.e., the vector $z$ --, with $n \gg k$ in almost all the practical cases. These equalities are in the form:
\begin{align}
    \label{eqn:fine_grained_constraints}
    \tilde{F_1}^T \tilde{F_j} \tilde\alpha_1^* = \tilde{F_j}^T \tilde{y} & & \forall j \in \{2, \hdots, k\}
\end{align}
where $\tilde{F}_j = x^j - \mu(x^j) = \tilde{x}^j$.

Since $\tilde\alpha_1^*$ can be computed in closed-form, we can plug it into \Cref{eqn:fine_grained_constraints}, eventually obtaining the following set of constraints:
\begin{align}
    \operatorname{cov}(x^j, x) \operatorname{cov}(x, y) = \operatorname{cov}(x^j, y) \operatorname{var}(x) & & \forall j \in \{2, \hdots, k\}
\end{align}
which can be used to solve \Cref{eqn:fine_grained_problem} without the need to set up the respective Least Squares Problems.

\subsection{Experimental Setup for Reproducibility}
\label{app:experimental_setup}
All the models are trained on a machine with an Intel Core I9 10920X 3.5G and 64GB of RAM.

The Random Forest and Gradient Boosting models are based on their available implementations in \texttt{scikit-learn 1.0.2} with default parameters, while the Neural Network and the semantics-based Regularization models are implemented using \texttt{torch 1.13.1}.
Specifically, the hyper-parameters of neural-based models are obtained via a grid search analysis with train-test splitting on the two unconstrained tasks aimed at maximizing test accuracy. 
In particular, the  neural models are trained for 200 epochs with batch size 128 and two layers of 256 units for the \textit{Communities \& Crimes} tasks and three layers of 32 units for the \textit{Adult} tasks. The only exception is the semantics-based Regularization model which runs for 500 epochs to compensate the fact that it is trained full-batch in order to better deal with group constraints.
Additionally, all the neurons have ReLU activation function except for the output one, which has either linear or sigmoid activation depending on whether the task is regression or classification, respectively.
Accordingly, the loss function is either mean square error or binary crossentropy, but in both cases the training is performed using the Adam optimizer with default hyperparameters.

As regards Moving Target's optimization routine, we leverage the Python APIs offered by \texttt{gurobipy 10.0} to solve it within our \texttt{Python 3.7} environment. The backend solver is \texttt{Gurobi 10.0}, for which we use the default parameters except for $\mbox{WorkLimit} = 60$.

\subsection{Constrained Approaches Descriptions}
\label{app:constrained_models}

Here we will briefly present the two approaches that we use to enforce our constraint in several ML models. Both approaches are employed to solve the following  constrained optimization problem:
\begin{align}
    \label{eqn:constrained_problem}
    \argmin_{\theta} \mathcal{L}(f(x; \theta), y) & & \text{s.t. } f(x; \theta) \in \mathcal{C}
\end{align}
where $f(x; \theta)$ represent the predictions of the ML model $f$ with learned parameters $\theta$, $\mathcal{C}$ is the feasible region, and $\mathcal{L}$ is a task-specific loss function.
For both the approaches, the original papers are provided as reference for a more details.

\paragraph{Moving Targets}
Moving Targets (MT)~\cite{Detassis_2021} is a framework for constrained ML based on bilevel decomposition.
It works by iteratively alternating between a \emph{learner step}, which is in charge of training the ML model, and a \emph{master step}, which projects the solution onto the feasible space while minimizing the distance between both model's predictions and original targets. In practice, MT solves the problem described in \Cref{eqn:constrained_problem} by alternating between the two following sub-problems:
\begin{align}
    \label{eqn:mt_master_step}
    z_{(i)} & = \argmin_z \mathcal{L}(z, f(x; \theta_{(i - 1)})) + \alpha_{(i)} \cdot \mathcal{L}(z, y)  & \text{s.t. } z \in \mathcal{C} \\
    \label{eqn:mt_learner_step}
    \theta_{(i)} & = \argmin_{\theta} \mathcal{L}(f(x; \theta), z_{(i)}) &
\end{align}
where the subscript ${(i)}$ indicates values obtained at the $i^{th}$ iteration, $\alpha_{(i)}$ is a factor used to balance the distance between the original targets and the predictions during the \emph{master step}, and the first value $\theta_{(0)}$ is obtained by pre-training the ML model.

The algorithm is perfectly suited for our purpose for three main reasons: 1) it is model-agnostic, thus allowing us to test the behaviour of our constraint for different models, each of which has its own specific bias and limitation; 2) it is conceived to deal with declarative group-constraints like the \ourmodel{} one, since it allows to train the ML model using mini-batches if needed; and 3) it can naturally deal with classification tasks without the need of relaxing the class targets to class probabilities.

As regards our experiments, the \textit{learner step} is performed as a plain ML task leveraging either \texttt{scikit-learn} or \texttt{torch} depending on the chosen model.
Instead, the \textit{master step} is formulated as a Mixed-integer Program (MIP) and delegated to the \texttt{Gurobi} solver.
More specifically, for the coarse-grained constraint formulation we define $k$ free variables for the coefficients $\tilde{\alpha}$ and retrieve their values by imposing an equality constraints according to \Cref{eqn:cst_set_1}; the overall constraint on the \ourmodel{} value is then imposed according to \Cref{eqn:cst_set_2}.
As for the fine-grained formulation, we force the constraint $-q \leq \operatorname{cov}(x, z) \leq q$ in order not to exceed the defined threshold and, additionally, we impose the satisfaction of the $k - 1$ equality constraints defined in \Cref{eqn:fine_grained_constraints}, as motivated by the optimized formulation showed in \Cref{app:fine_grain_formulation}.

Finally, in our setup we define each value $\alpha_{(i)}$ as the $i^{th}$ item of the harmonic series, namely $\alpha_{(i)} = i^{-1}$, and the loss function $\mathcal{L}$ is either MSE or Hamming Distance depending on whether the task is regression or classification.
Respectively to the Semantic-based Regularization approach, a key advantage of MT lies in the fact that MIP models can naturally deal with discrete variables, thus requiring no need to relax the problem to the continuous domain.

\paragraph{Semantic-based Regularization}
The Lagrangian Dual framework for Semantic-based Regularization~\cite{Fioretto2021} extends the concept of loss penalizers by allowing for an automated calibration of the lagrangian multipliers.
Specifically, let us consider the case in which we have a penalty vector $\mathcal{P}(y, f(x; \theta)) \in \mathbb{R}^{+k}$ which represents the violations for $k$ different constraints.
We can embed these violations in the loss function $\mathcal{L}(y, f(x; \theta)) \in \mathbb{R}^+$ of our neural model by multiplying each violation with its respective multiplier $\lambda_i$.
The overall loss will be 
$$\mathcal{L}(y, f(x; \theta)) + \lambda^T \mathcal{P}(y, f(x; \theta)).$$
The main pitfall of this approach is that it requires to fine-tune the multipliers vector $\lambda$ according to the task. The Lagrangian Dual framework solves this problem by proposing a bilevel optimization schema where: 1) the loss function is minimized via gradient-descent with fixed multipliers, and 2) the loss function is maximized via gradient-ascent with fixed network structure. In practice, this is equivalent to perform the following steps in sequence:
\begin{align}
    \theta_{(i)} & = \argmin_{\theta} \left\{\mathcal{L}(y, f(x; \theta)) + \lambda_{(i-1)}^T \mathcal{P}(y, p) \right\} \\
    \lambda_{(i)} &= \argmin_{\lambda} \left\{ \mathcal{L}(y, f(x; \theta_{(i)})) + \lambda^T \mathcal{P}(y, p) \right\}
\end{align}
where the subscript ${(i)}$ indicates the value of the $i^{th}$ iteration -- performed once per batch --, and $\lambda_{(0)}$ is a null vector.

As regards our experiments, the $\tilde{\alpha}$ coefficients are computed via the \texttt{torch.linalg.lstsq} differentiable operator. In the coarse-grained formulation, the penalizers vector $\mathcal{P}(y, f(x; \theta))$ consists of a single which is computed according to \Cref{eqn:gedi_vector_penalizer}. Instead, in the fine-grained formulation the vector has $k$ different terms -- one for each $\tilde{\alpha}_i$ --, of which all but the first term exhibit a violation proportional to their absolute value.

A major pitfall of this approach is due to the incompatibility of the \textit{round} operator with gradient-based learning algorithms, being its gradient null for each $x$. This makes it necessary to relax the formulation of the \ourmodel{} indicator (only) in classification tasks, by adopting predicted probabilities rather than predicted class targets.